# A Simple Haploid-Diploid Evolutionary Algorithm

Larry Bull

Computer Science Research Centre

University of the West of England, Bristol, UK

larry.bull@uwe.ac.uk

**Abstract**

It has recently been suggested that evolution exploits a form of fitness landscape smoothing within eukaryotic sex due to the haploid-diploid cycle. This short paper presents a simple modification to the standard evolutionary computing algorithm to similarly benefit from the process. Using the well-known NK model of fitness landscapes it is shown that the benefit emerges as ruggedness is increased.

# Introduction

Simple bacteria are haploid and contain one set of genes, whereas the more complex eukaryotic organisms – such as plants and animals - are predominantly diploid and contain two sets of genes. It has recently been suggested that natural evolution exploits a form of generalization with eukaryotes under which the fitness landscape is smoothed [Bull, 2017]. The vast majority of work within evolutionary computation has used a haploid representation scheme, ie, individuals are one solution to the given problem. However, a small body of work exists using a diploid representation scheme, ie, individuals carry two solutions to the given problem. In such cases recombination typically occurs between corresponding haploids/genes in each parent, essentially doubling the standard process, and a dominance scheme is utilized to reduce the diploid down to a traditional haploid solution for evaluation. That is, as individuals carry two sets of genes/variables, a heuristic is included to choose which of the genes to use (eg, see [Bhasin & Mehta, 2016] for a recent review).

Eukaryotes exploit a haploid-diploid cycle where haploid cells are brought together to form the diploid cell/organism. At the point of reproduction by the cell/organism, the haploid genomes within the diploid each form haploid gamete cells that (may) join with a haploid gamete from another cell/organism to form a diploid (Figure 1). Specifically, each of the two genomes in an organism is replicated, with one copy of each genome being crossed over. In this way copies of the original pair of genomes may be passed on, mutations aside, along with two versions containing a mixture of genes from each. Previous explanations for the emergence of the alternation between the haploid and diploid states are typically based upon its being driven by changes in the environment (after [Margulis & Sagan, 1986]). Recently, an explanation for the haploid-diploid cycle in eukaryotes has been presented [Bull, 2017] which also explained other aspects of their sexual reproduction, including the use of recombination, based upon the Baldwin effect [Baldwin, 1896].

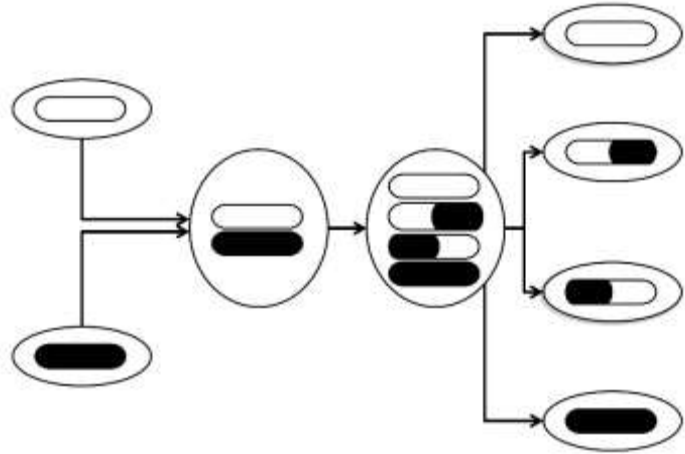

Figure 1: Two-step meiosis with recombination under haploid-diploid reproduction as seen in most eukaryotic organisms (after [Maynard Smith & Szathmary, 1995]).

Key to the new explanation for the evolution of eukaryotes is to view the process from the perspective of the constituent haploids: *a diploid organism may be seen to simultaneously represent two points in the underlying haploid fitness landscape.* The fitness associated with those two haploids is therefore that achieved in their combined form as a diploid; each haploid genome will have the same fitness value and that will almost certainly differ from that of their corresponding haploid organism due to the interactions between the two genomes. That is, the effects of haploid genome combination into a diploid can be seen as a simple form of phenotypic plasticity for the individual haploids before they revert to a solitary state during reproduction. In this way evolution can be seen to be both assigning a single fitness value to the *region* of the landscape between the two points represented by a diploid's constituent haploid genomes and altering the shape of the haploid fitness landscape. In particular, the latter enables the landscape to be smoothed under a rudimentary Baldwin effect process [Hinton & Nowlan, 1987], whilst the former can be seen to represent a simple form of generalization over the landscape.

Numerous explanations exist for the benefits of recombination (eg, [Bernstein and Bernstein, 2010]) but the role becomes clear under the new view: recombination facilitates genetic assimilation within the simple form of the Baldwin effect. If the haploid pairing is beneficial and the diploid is chosen under selection to reproduce, the recombination process can bring an assortment of those partnered genes together into new haploid genomes. In this way the fitter allele values from the pair of partnered haploids may come to exist within individual haploids more quickly than the under mutation alone (see [Bull, 2017] for full details).

This aspect of eukaryote reproduction has recently been applied directly to the standard evolutionary computing algorithm [Bull, 2017b]. That is, each individual was a diploid and had both sets of genes evaluated on the given objective function. The effects on fitness from their combination within a single organism/solution was captured through simple averaging (after [Bull, 2017]). Thus in comparison to the standard evolutionary algorithm using haploids, each generation required twice as many evaluations for the same population size. The rest of the reproduction was as in Figure 1. Results suggested an increase in performance over haploids on more rugged fitness landscapes, with the latter allowed twice as many generations as the former for equivalence.

This paper presents a simplified haploid-diploid evolutionary algorithm which exploits the same generalization process over the fitness landscape but maintains a traditional population of haploid solutions which are evaluated in the standard way. The NK model is used to explore relative performance.

**The NK Model**

Kauffman and Levin [1987] introduced the NK model to allow the systematic study of various aspects of fitness landscapes (see [Kauffman, 1993] for an overview). In the standard model, the features of the fitness landscapes are specified by two parameters: *N*, the length of the genome; and *K*, the number of genes that has an effect on the fitness contribution of each (binary) gene. Thus increasing *K* with respect to *N* increases the epistatic linkage, increasing the ruggedness of the fitness landscape. The increase in epistasis increases the number of optima, increases the steepness of their sides, and decreases their correlation. The model assumes all intragenome interactions are so complex that it is only appropriate to assign random values to their effects on fitness. Therefore for each of the possible K interactions a table of $2^{(K+1)}$ fitnesses is created for each gene with all entries in the range 0.0 to 1.0, such that there is one fitness for each combination of traits (Figure 2). The fitness contribution of each gene is found from its table. These fitnesses are then summed and normalized by N to give the selective fitness of the total genome. All results reported in this paper are the average of 10 runs (random start points) on each of 10 NK functions, ie, 100 runs, for 20,000 generations. Here $0 \leq K \leq 15$, for N=50 and N=100.

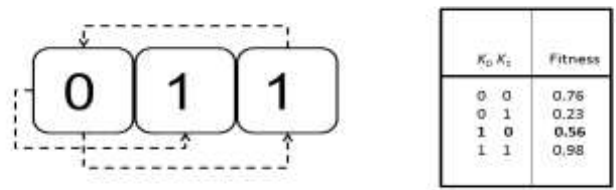

Figure 2: An example NK model (N=3, K=1) showing how the fitness contribution of each gene depends on K random genes (left). Therefore there are $2^{(K+1)}$ possible allele combinations per gene, each of which is assigned a random fitness. Each gene of the genome has such a table created for it (right, centre gene shown). Total fitness is the normalized sum of these values.

**A Haploid Diploid Evolutionary Algorithm**

Figure 3 (top) shows a schematic of a traditional evolutionary algorithm (EA) which exploits one-point recombination, single-point mutation, and creates one offspring per cycle (steady state) which replaces the worst individual in the population here. Figure 3 (bottom) shows how the generalization mechanism described above is implemented on top of that process. As can be seen: a traditional population of haploid individuals is maintained (A); a temporary population of diploid solutions is created from them by copying each haploid individual and then another haploid is chosen at random

(B), with the fitness of the two haploids averaged (C); binary tournament selection then uses those fitnesses to pick two diploid parents (D); the haploid-diploid reproduction cycle with two-step meiosis as shown in Figure 1 is then used for the two chosen parents (E); one of the resulting haploids is chosen at random, mutated, and evaluated (F); and, the offspring haploid is inserted into the original population (G).

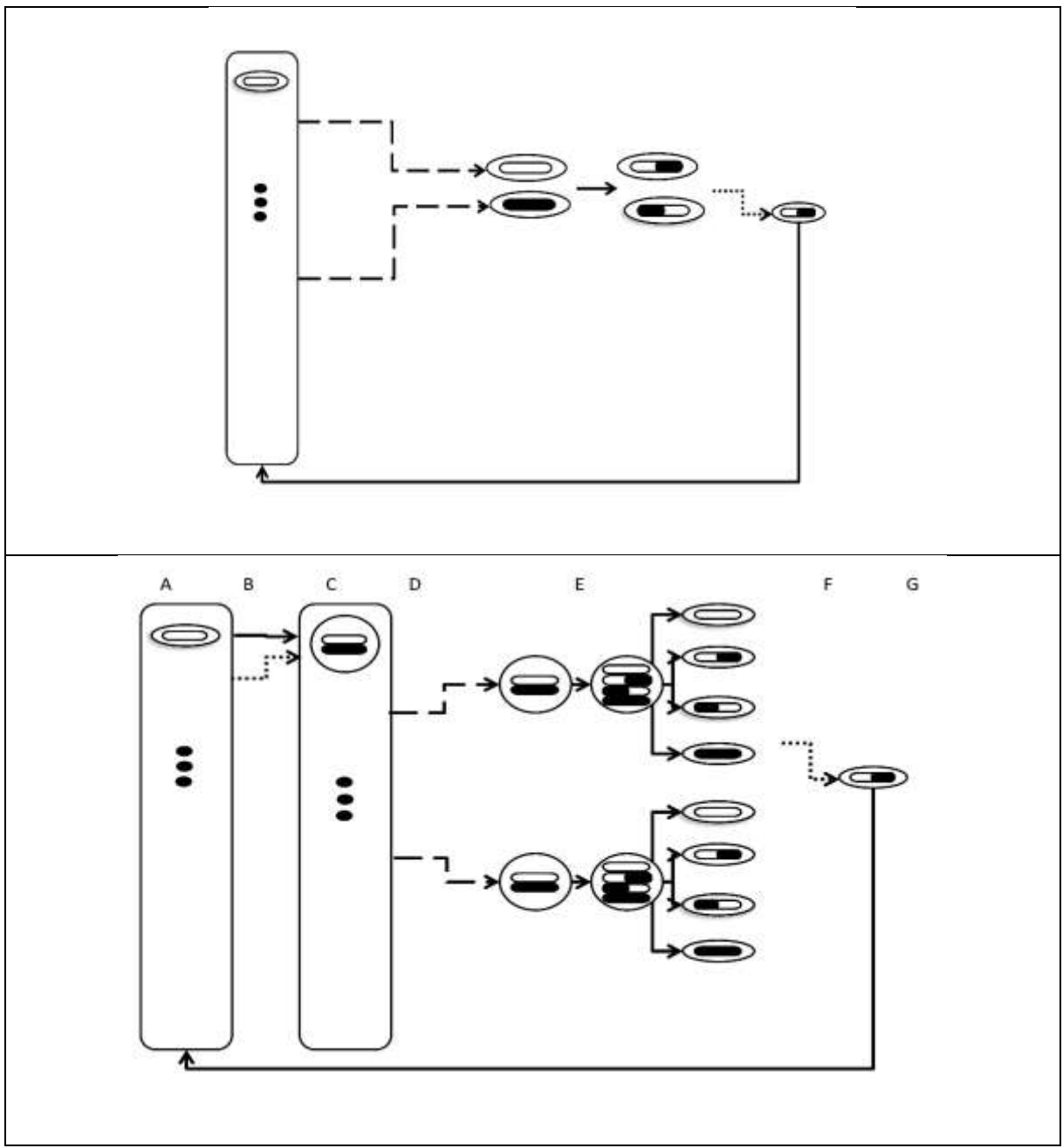

Figure 3: A schematic of the traditional evolutionary algorithm (top) and of the simple haploid-diploid evolutionary algorithm explored here (bottom).

Figure 4 shows example results from running both the standard EA and the haploid-diploid EA (HDEA) on various NK fitness landscapes. Here population size $P$=30. As can be seen, when $K$>4, the HDEA performs best for *N=50 and K>2 for N*=100 (T-test, $p$<0.05). Thus, as anticipated, the generalization process proves beneficial with increased fitness landscape ruggedness. Figure 5 shows examples of how this is also true for different $P$, although the benefit is lost for higher $K$ when $P$=10. Related to this, since the HDEA makes a temporary population of diploids containing extra copies of randomly chosen haploid solutions, it might be argued that a larger population is available to selection than in the standard EA. Moreover, as the underlying traditional haploid population converges upon higher fitness solutions, the random sampling could be increasing their number and thereby altering the comparative selection pressure over time. However, results from simply creating a temporary haploid population of size 2$P$ in the same way as the diploid population does not alter performance significantly (not shown, eg, see [Karafotias et al, 2015] for discussions of dynamic population sizing in general).

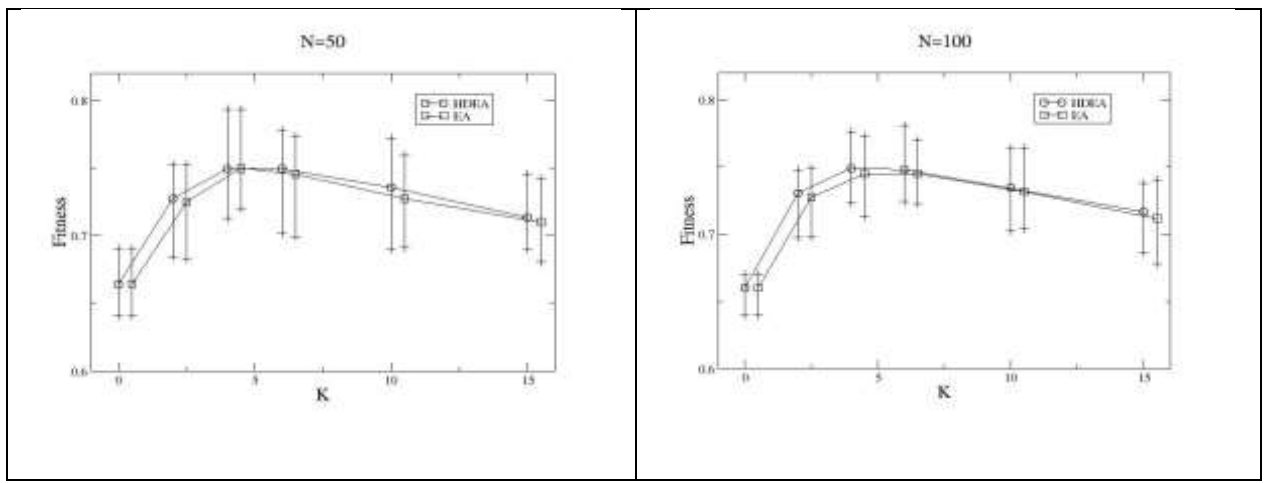

Figure 4: Showing examples of the fitness reached after 20,000 generations on landscapes of various size ($N$) and ruggedness ($K$). Error bars show min and max values.

## Conclusion

In the standard evolutionary computing approach each individual solution can be seen to represent a single point in the fitness landscape. The same is true of bacteria in natural evolution. It has recently been suggested that natural evolution is using a more sophisticated approach with eukaryotes, exploiting a generalization process whereby each individual represents a region in the fitness landscape [Bull, 2017]. Of course, landscape smoothing can be achieved by numerous mechanisms (after [Hinton & Nowlan, 1987]) but they all require extra fitness evaluations. The scheme presented in this paper is intended to obtain the same effect through what is essentially simple population manipulation rather than through altering the underlying representation and evaluations of the standard evolutionary computing approach. Benefits are seen on more rugged fitness landscapes, as anticipated by [Bull, 2017]. Future work will explore the utility of the approach on tasks where fitness evaluations are computationally expensive, in particular.

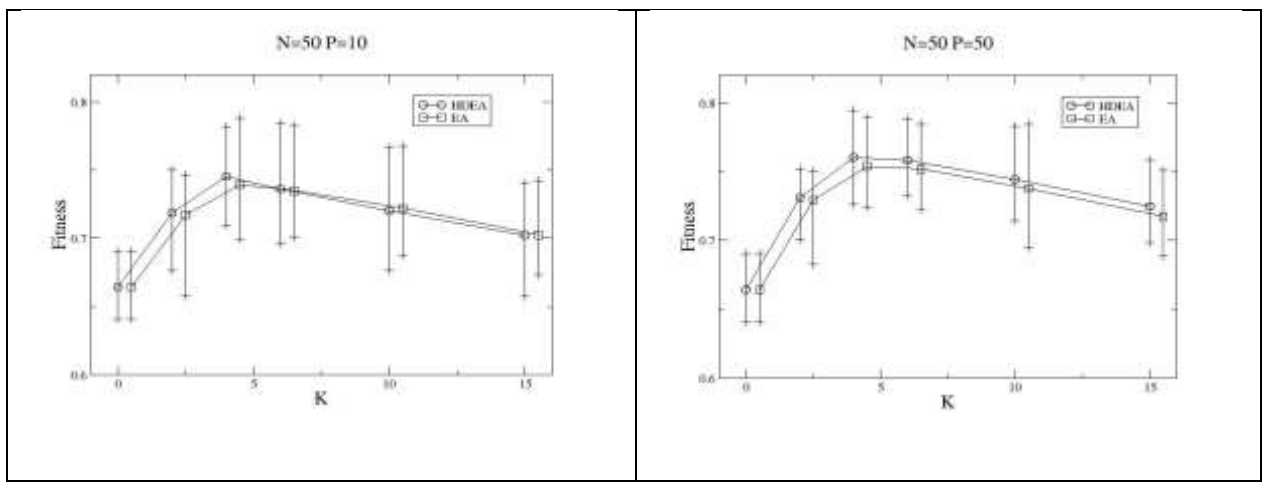

Figure 5: Showing examples of the fitness reached after 20,000 generations with differing population sizes (*P*).

**Acknowledgement**

This work was supported by the European Research Council under the European Union's Horizon 2020 research and innovation programme under grant agreement No. 800983.

**References**

Baldwin, J. M. (1896) A new factor in evolution. *American Naturalist* 30: 441–451.

Bernstein, H. & Bernstein, C. (2010) Evolutionary origin of recombination during meiosis. *BioScience* 60: 498-505

Bhasin, H. & Mehta, S. (2016) On the applicability of diploid genetic algorithms. *AI & Society* 31(2): 265-274.

Bull, L. (2017) The evolution of sex through the Baldwin effect. *Artificial Life* 23(4):481-49

Bull, L. (2017b) Haploid-Diploid Evolutionary Algorithms: the Baldwin Effect and Recombination Nature's Way. In *Proceedings of the 2017 AISB Convention.* AISB. Available arXiv:1607.00318: https://arxiv.org/ftp/arxiv/papers/1608/1608.05578.pdf

Hinton, G. E., & Nowlan, S. J. (1987) How learning can guide evolution. *Complex Systems* 1:495–502

Karafotias, G., Hoogendoorn, M. & Eiben, A.E. (2015) Parameter control in evolutionary algorithms: Trends and challenges. *IEEE Transactions on Evolutionary Computation* 19:167{187

Kauffman, S. A. (1993) *The Origins of Order: Self-organisation and Selection in Evolution*. New York, NY: Oxford University Press.

Kauffman, S. A., & Levin, S. (1987) Towards a general theory of adaptive walks on rugged landscapes. *Journal of Theoretical Biology* 128: 11–45

Margulis, L. & Sagan, D. (1986) *Origins of Sex: Three Billion Years Recombination*. Yale University Press, New Haven.